\documentclass[runningheads,a4paper]{llncs}

\RequirePackage{fix-cm}

\usepackage{amssymb}
\usepackage{graphicx}
\usepackage{amsmath}
\usepackage{mathtools}

\newcommand{\E}{{\mathbb{E}}}

\newcommand{\D}{{\mathcal{D}}}
\newcommand{\N}{{\mathcal{N}}}
\newcommand{\auxR}{{\mathcal{R}}}
\newcommand{\h}{{\bf h}}
\newcommand{\x}{{\bf x}}
\newcommand{\y}{{\bf y}}
\newcommand{\auxL}{{\mathcal{L}}}
\newcommand{\Diverg}[3]{#1\,(#2,\;#3)}

\usepackage{url}
\newcommand{\keywords}[1]{\par\addvspace\baselineskip
\noindent\keywordname\enspace\ignorespaces#1}

\begin{document}

\mainmatter  

\title{Guided Layer-wise Learning for Deep Models \\ using Side Information}

\titlerunning{Guided Layer-wise Learning}

%
%
\author{Pavel Sulimov, Elena Sukmanova, \\Roman Chereshnev, and Attila Kert\'esz-Farkas}


\institute{Department of Data Analysis and Artificial Intelligence, Faculty of Computer Science, \\ National Research University Higher School of Economics (HSE) \\ 
	3 Kochnovsky Proezd, Moscow, Russian Federation\\
	Corresponding Author: Attila Kertesz-Farkas ORCID: 0000-0001-8110-7253
	\email{akerteszfarkas@hse.ru}}

\authorrunning{P. Sulimov et al.}


%
%

\maketitle
\vspace{-0.5cm}

\begin{abstract}
Training of deep models for classification tasks is hindered by local minima problems and vanishing gradients, while unsupervised layer-wise pretraining does not exploit information from class labels. Here, we propose a new regularization technique, called diversifying regularization (DR), which applies a penalty on hidden units at any layer if they obtain similar features for different types of data. For generative models, DR is defined as divergence over the variational posteriori distributions and included in the maximum likelihood estimation as a prior. Thus, DR includes class label information for greedy pretraining of deep belief networks which result in a better weight initialization for fine-tuning methods. On the other hand, for discriminative training of deep neural networks, DR is defined as a distance over the features and included in the learning objective. With our experimental tests, we show that DR can help the backpropagation to cope with vanishing gradient problems and to provide faster convergence and smaller generalization errors.
\end{abstract}
\vspace{-1cm}
\section{Introduction}
\label{intro}
Deep models \cite{bengio2009learning}, especially deep neural networks (DNNs), iteratively process data through various abstraction levels as $\x=\h_0\rightarrow \h_1 \rightarrow \h_2 \rightarrow \dots \rightarrow \h_L =y$. The first layer $\x$ is the raw data layer, and higher layers $\h_l$ aim to give a higher abstraction of the data, often referred to as features. The last layer can correspond either to class labels $y$ in classification tasks or some other high-level cause in generative tasks. Each layer utilizes a monotone, non-linear so-called activation function $g_l(\h_l^T\theta_l)\rightarrow \h_{l+1}$ to transform features $\h_{l}$ to $\h_{l+1}$, where $\theta_l$ denotes the parameterization of the feature transformation at the given layer.

Currently, there are two main approaches for training deep models: convolutional and non-convolutional. In the non-convolutional approach, the data distribution is modeled by generative probabilistic distributions via maximizing the likelihood. One of the standard approaches is an unsupervised layer-wise pretraining \cite{bengio2007greedy,salakhutdinov2009deep,larochelle2009exploring,arnold2012layerwise,salakhutdinov2010efficient}. This is a bottom-up approach, in which a deep model is built via iteratively stacking single-layer generative models until the desired depth is achieved, and the procedure is finalized by using either backpropagation for discriminative fine-tuning or, e.g., a wake-sleep algorithm for generative fine-tuning \cite{Goodfellow-et-al-2016}. In each step, a generative model is often trained to model either a joint distribution  $P(\h_l, \h_{l+1}; \theta_l)$ using restricted Boltzmann machines (RBMs) or $P(\h_{l+1}|\h_l)$ using sigmoid belief networks (SBNs), where $\h_l$ represents the input data obtained from the previous layer, and $\h_{l+1}$ is treated as latent variable.  Then, the subsequent layer is to learn a distribution of data obtained by sampling the hidden units of the previous layer. For the sake of simplicity, let us denote the input data by $\x$ and the latent variables by $\h$ at any layers. For a given set of data $\D=\{\x_i\}$, the model parameters are estimated by maximizing the data log likelihood, which is given as: 

\begin{equation}
l(\theta;\D)=\log P(\D;\theta) =\sum_{\x_i \in \D} \log \sum_\h P(\x_i,\h;\theta). 
\label{eq:loglikelihood}
\end{equation}


The optimization of Eq. \ref{eq:loglikelihood}
w.r.t. $\theta$ is often done via the Markov chain Monte Carlo (MCMC) technique or via variational methods. The former tends to suffer from slow mixing rates and is too computationally expensive to be practical \cite{Neal1992,Jordan2003mcmc}. Variational methods \cite{Jordan1999variational} utilize an auxiliary so-called variational distribution $Q_\x(H;\phi)$ over the hidden variables for every given data $\x$ in order to approximate $P(H|\x;\theta)$ wherein the approximation is solved via optimization. Here, we introduce  $Q_\x^\phi$ and $P_\x^\theta$ to shorten $Q(H|x;\phi)$ and $P(H|x;\theta)$, respectively. Then the log likelihood can be bounded from below as:
\begin{equation}
\begin{aligned}
l(\theta;\D) & \ge \sum_{\x \in \D}\sum_\h{Q_\x(\h;\phi)\log\frac{P(\x,\h;\theta)}{Q_\x(\h;\phi)}} \\
&=\sum_{\x \in \D}\left(\E_{Q_\x}[\log P(\x,\h;\theta)]-\E_{Q_\x}[\log Q_\x^\phi]\right) & =\auxL(\theta,\phi;\D).& 
\label{eq:loglikelihood_bound}
\end{aligned} 
\end{equation}
The last term $\auxL$ is a lower bound on the original log likelihood and can be decomposed into two parts as:
\begin{equation}
\auxL(\theta,\phi;\D)=\sum_{\x\in\D}\log P(\x;\theta)- \sum_{\x\in\D}\Diverg{KL}{Q_\x^\phi}{P_\x^\theta}.
\label{eq:loglikelihood_bound1}
\end{equation}

This means that a tight lower bound can be achieved by minimizing the Kullback-Leibler (KL) divergence between the variational distribution $Q$ and the exact posterior distribution $P$. 
Following \cite{Jordan1999variational,csisz1967information} in variational perspectives, the parameters maximizing the lower bound of the log likelihood $\auxL$ can be obtained by iteratively alternating between the following two optimization steps:\\
\indent\; {\bf E-step}: $\phi^{(t+1)} \leftarrow \text{argmax}_{\phi\in M}\;{\auxL(\theta^{(t)},\phi;\D)}$\\
\indent\; {\bf M-step}: $\theta^{(t+1)} \leftarrow \text{argmax}_{\theta\in \Omega}\;{\auxL(\theta,\phi^{(t+1)};\D)}$\\
\noindent until convergence, where $M$ and $\Omega$ are the corresponding parameter spaces.

In the convolutional approach, deep neural network models are made of convolutional layers, rectified linear units, and pooling layers \cite{lecun1998gradient,Goodfellow-et-al-2016,lecun1989generalization} inspired by the mammal's visual cortex \cite{hubel1962receptive,fukushima1982neocognitron}, and they are referred to as  convolutional neural networks (CNNs). CNNs are said to yield better gradients for the backpropagation algorithm in contrast to fully connected, deep, sigmoid neural networks. As a result, CNNs usually provide improved performance on visual-related or speech recognition-related tasks. 

In both approaches, convolutional and non-convolutional, the training of $\theta_l$ parameters in deep models is notoriously hard, and it is often viewed as an art rather than a science. There are four main problems with training deep models for classification tasks: 
(i) Training of deep generative models via an unsupervised layer-wise manner does not utilize class labels, therefore essential information might be neglected. 
(ii)  When a generative model is learned, it is difficult to track the training, especially at higher levels \cite{Goodfellow-et-al-2016}. For DNNs, the backpropagation method suffers from a problem known as vanishing gradients \cite{erhan2009difficulty}.   
(iii)  In principle, a generative model can be fitted to data arbitrarily well \cite{Sutskever2008DeepNS,hornik1991approximation,hartman1990layered}, in practice, the optimization procedure with latent variables can stuck in a poor local minima.	(iv) The structure of the model is often specified in advance, and the designed model might not fit the data well. In particular, the number of hidden units or layers is often defined by the experimenter's intuition or habits; however, it is hard to give a bone fide estimation on the numbers of the latent components. 

Here, we introduce a new regularization method, called diversifying regularization (DR), on the hidden units for training deep  models for classification tasks. In principle, the proposed regularizer favors different abstract representation for two data samples belonging to different classes. This regularization is denoted by $D(\h_p^{(l)},\h_q^{(l)})$, where $\h_p^{(l)}$ and $\h_q^{(l)}$ are abstract representations of data $\x_p$ and $\x_q$ (resp.)  at layer $l$, and the data are of different types $(y_p\ne y_q)$.  
For maximum likelihood estimation of generative models, we define DR in terms of divergence function $D(Q_{x_p},Q_{x_q})$ and include it in Eq.  \ref{eq:loglikelihood_bound1} as an additive term. For discriminative learning of DNNs using backpropagation, we introduce DR as a distance function and include it in the learning objective as an additive cost. 

We anticipate that DR helps cope with the aforementioned four problems. (i) DR is constructed based on class labels, and it can be employed on any abstract representation pairs at any hidden layer. Therefore, it implicitly includes information about the data classes in the unsupervised layer-wise pretraining. 
(ii) As a consequence of (i), DR can guide the pretraining at higher levels toward solution which obtain good classification performance.
(iii) DR can help direct the gradient ascent optimizer toward an optimum where the internal representation of different types of data is more different.  This could be particularly significant in the early steps in the optimization procedure when the gradients could be directed toward regions containing such solutions.  (iv) Good regularization techniques can mitigate the impact of model structure construction problems, and they should help cope with overfitting problems and provide smaller generalization errors over a wider range of structures.


It could also be possible to reward hidden units if they learn similar abstract representations for data of the same type. In our opinion, this hampers the model's ability to discover fine subgroups in the data. For instance, considering a model that distinguishes between, e.g., cats and dogs, we might not want to force the model to learn similar representations for tigers and Persian cats, especially at lower abstraction levels.

There have been several attempts to incorporate class labels into RBMs. For instance, Larochelle et al. \cite{larochelle2008classification} coupled a data vector with the one-hot-encoded class labels as $(\x,\y)$ and trained a RBM. The drawback of this representation is that it requires significantly more weights, which linearly depend on the number of the classes.

This article is organized as follows. In the next section, we introduce DR for probabilistic generative models, where  particular emphasis is placed on RBMs and Variational Autoencoders (VAEs). In section 3, we introduce DR for discriminative DNN. In section 4, we present and discuss experimental results, and finally, we summarize our conclusions in the last section.

\section{Diversifying Regularization Based on Side Information for Generative Models}

Let us introduce a regularization method for latent variable generative models using side information. In general, side information refers to knowledge that is neither in the input data nor in the output label space but includes useful information for learning. The idea of side information was introduced in \cite{xing2002distance} and applied recently in \cite{JonschkowskiHB15}. Let $\D=\{(\x_i, y_i)\}$ be a given labeled training set. Using  class labels from the training data, we construct side information by defining a set of data pairs, where the members of each pair belong to different classes, that is, if $(\x_p,\x_q)\in \N$ then $y_p\ne y_q$.  We note that the size of $\N$ can be very large in the case of big data; however, we think that it is not necessary to include all possible pairs but just a smaller subset or only pairs of data that are close to each other in terms of some distance measure. 

We define the regularization term, denoted by $D$, by divergence functions over the distribution functions over the latent variables. For instance, for two given data $\x_p$ and $\x_q$ the regularization is defined as $ \Diverg{D}{Q_{\x_p}^\phi}{Q_{\x_q}^\phi} $. This term can be included to the MLE as follows:
\begin{equation}
\begin{aligned}
r(\theta,\phi;\D) = \sum_{\mathclap{\x\in\D}} \log \sum_\h P(\x,\h;\theta)  
+\alpha\sum_{\mathclap{(\x_p,\x_q)\in \N}} \Diverg{D}{Q_{\x_p}^\phi}{Q_{\x_q}^\phi},
\label{eq:MLE}
\end{aligned} 
\end{equation}
where $r$ stands for regularized likelihood. The second term introduces a penalty on two distributions if they are similar, but they should not be, and $\alpha$ is a trade-off parameter. $D$ denotes a divergence function for probability distributions. A good source of divergence functions is provided by, e.g., Csisz\'ar's f-divergence class \cite{csisz1967information}. It is defined as follows: 

\begin{definition} Given any continuous, convex function $f:[0,+\infty)\rightarrow R \cup \{+\infty\}$, such that $f(1)=0$, then the f-divergence between distributions $P$ and $Q$ is measured as
	\begin{equation}
	\Diverg{D_f}{P}{Q}
	=\sum_z q_zf\left(\frac{p_z}{q_z}\right).
	\label{eq:f-div}
	\end{equation}
\end{definition}
The Kullback-Leibler divergence is an $f$-divergence, if $f(t)=t\log t$. We were interested in using a symmetric divergence measure; therefore, we decided to chose the Hellinger divergence. It is defined as $\Diverg{D_H}{P}{Q}   =1-\sum_z\sqrt{p_zq_z}$ generated by $f(t)=1-\sqrt{t}$. We note that the log of the Hellinger divergence is closely related to the Bhattacharyya distance (BD) (also known as a R\'enyi divergence at $\alpha=0.5$), which is defined as $D_B(P\;\|\;Q)=-\ln\sum_z\sqrt{p_zq_z}$. BD has been successfully used in a wide range of applications, including feature extraction and selection \cite{choi2003feature}, image processing \cite{goudail2004bhattacharyya}, and speaker recognition \cite{you2009svm}.


The lower bound of $r$ can be obtained again as
\begin{equation}
\begin{aligned}
\auxR&(\phi, \theta;\D) = \sum_{\x \in \D} \log P(\x;\theta) 
+ \alpha \sum_{\mathclap{(\x_p,\x_q)\in \N}} D_H(Q_{\x_p}^\phi,Q_{\x_q}^\phi) 
- \sum_{\x\in\D}\Diverg{KL}{Q_\x^\phi}{P_\x^\theta}.
\end{aligned}
\label{eq:loglikelihoodsideinfo}
\end{equation}


The optimization of Eq. \ref{eq:loglikelihoodsideinfo} can be carried out with, for instance, iterative gradient-based methods. By using the EM algorithm, we get:\\
\indent \; {\bf E-step}: 
$\phi^{(t+1)} \leftarrow \text{argmax}_{\phi\in M}\;\left\{ \; \alpha\sum\limits_{\mathclap{(\x_p,\x_q)\in \N}} D_H(Q_{\x_p}^\phi,Q_{\x_q}^\phi) \right. 
\left. - \sum_\x\Diverg{KL}{Q_\x^\phi}{P_\x^{\theta^{(t)}}}  \right\}$

\indent \; {\bf M-step}: $\theta^{(t+1)} \leftarrow \text{argmax}_{\theta\in\Omega}\;{\sum_{i} \log  P(\x_i;\theta)}.$

The derivatives of the regularization $D$ can be calculated analytically; however, they can be easily calculated automatically using recent toolboxes, e.g., with Theano's grad function.

\subsection{Restricted Boltzmann Machines (RBMs) Using Side Information}

Restricted Boltzmann machines are particular type of Markov random fields that have two types of random binary variables, visible and hidden, denoted by $\x$ and $\h$ respectively. The visible and hidden units are connected, but there are no connections among the visible units and among the hidden units.
The probability of a given input in this model parameterized by $\theta$ is: 
\begin{equation}
P(\x;\theta)=\frac{1}{Z(\theta)}\sum_{\h}\exp(-E_\theta(\x,\h)),
\end{equation}
where $E_\theta(\x,\h)=-\sum_{ij}x_i\theta_{ij}h_j$ is the energy, $\theta_{ij}$s are the weights between visible unit $x_i$ and hidden unit $h_j$,  and $Z(\theta)$ denotes the partition function. Here, we omit the feature weights for $\x$ and $\h$ for the sake of simplicity; however, they can be included in the model and in our diversifying regularization.  The conditional distributions over hidden units $\h$ and visible units $\x$ can be formulated by logistic functions:
\begin{equation}
\begin{aligned}
P(\h|\x;\theta)  = \prod_j P(h_j|\x;\theta),  \quad \text{where }\; p(h_j=1|\x;\theta) = \sigma(\sum_i\theta_{ij}x_i) \\
P(\x|\h;\theta)  = \prod_i P(x_i|\h;\theta), \quad \text{where }\; p(x_j=1|\h;\theta) = \sigma(\sum_j\theta_{ij}h_j),
\end{aligned}
\end{equation}
where $\sigma(x)=(1+\exp(-x))^{-1}$ is the logistic function. The derivatives of the log-likelihood $l(\theta;\D)$ w.r.t. the model parameters using a set of data are: 
\begin{equation}
\frac{\partial l(\theta;\D)}{\theta_{ij}} = \sum_{\x_k}\sum_{\h}P(\h|\x_k;\theta){x_k}_ih_j - \sum_{\x,\h}P(x,h)x_ih_j, 
\label{eq:rmb_update}
\end{equation}
where $\x,\h$ runs over all possible states. The first term is called the data-dependent term, and Salakhutdinov et al \cite{salakhutdinov2010efficient} proposed a mean-field algorithm to approximate the posterior distribution $Q(\h|\x;\phi)$. The posterior distribution fully factorizes and can be written in the form $Q_\x(H)=\prod_j Q_\x^j(H_j)$, where $Q_\x^j(H_j) = (\sigma(z_x^j))^{H_j}(1-\sigma(z_x^j))^{1-H_j}$ and $z_x^j= \sum_i\theta_{ij}x_i$. Because of the factorization property, the mean filed algorithm can provide a fixed-point equations
$\mu^j_x=\sigma(z_x^j+\sum_{i\ne j}\mu^i_x)\quad (\forall j)$
and, therefore, the posteriors are  approximated by $Q_x^{j\,\phi}(H=1)=\mu^j_x$. It has been shown that mean field approximation provides a fast estimation for the posterior in practice. For more details, we refer the reader to \cite{salakhutdinov2010efficient}. The second term of Eq. \ref{eq:rmb_update} is called the data-independent term, and it is typically approximated by stochastic MCMC sampling methods \cite{Neal1992}, such as the $k$-step contrastive divergence (CD) or persistent CD algorithms \cite{tieleman2008training,yuille2006convergence}.

Now, we introduce the diversifying regularization for the variational optimization. The regularization is introduced on the individual factors $Q_x^j$  in the following way:
\begin{equation}
\begin{aligned}
D_H&(Q_{\x_p}^\phi,Q_{\x_q}^\phi) = 1-\sum_{\mathclap{h=\{0,1\}}}\sqrt{Q_{\x_p}^{j\,\phi}(h)Q_{\x_q}^{j\,\phi}(h)}
&= 1-\sqrt{\overline{\sigma}(z_p^j)\overline{\sigma}(z_q^j)} - \sqrt{\sigma(z_p^j)\sigma(z_q^j)},
\end{aligned}
\label{eq:RBMDR}
\end{equation}
where $\overline{\sigma}(z) = 1 -\sigma(z)$. The derivation of the regularizer's gradient can be obtained as follows:
\begin{equation}
\begin{split}
&\frac{\partial}{\phi_{ij}} D_H(Q_{\x_p}^\phi,Q_{\x_q}^\phi)  
=\sum_{\mathclap{h=\{0,1\}}}(-1)^h\frac{1}{2}\sqrt{Q_{\x_p}^{j\,\phi}(h)Q_{\x_q}^{j\,\phi}(h)}[Q_{\x_p}^{j\,\phi}(\overline{h}){x_p}_i+Q_{\x_q}^{j\,\phi}(\overline{h}){x_q}_i],
\label{eq:drboltzmann}
\end{split}
\end{equation}
where $\overline{h} = 1-h$. 

We introduce the total diversifying regularization to the data log likelihood as follows: 
\begin{equation}
r(\theta;\D) = \sum_{x\in \D}\log P(\x) +  \alpha \sum_{\mathclap{(\x_p,\x_q)\in \N}} D_H(Q_{\x_p}^\phi,Q_{\x_q}^\phi).
\end{equation}
Now, the weight update rule for the data-dependent term using mean filed approximation, and for the data-independent term using stochastic approximation, we get:
\begin{equation}
\begin{aligned}
\frac{\partial r(\theta;\D)}{\theta_{ij}} &=  \sum_{\x_l}{x_l}_i\mu^j_{x_l} - x_i^{(k)}h_j^{(k)} +\\ +&\;\alpha\sum_{{(\x_p,\x_q)\in  \N}}\left(\sqrt{\overline{\mu}^j_{x_p}\overline{\mu}^j_{x_q}}[\mu^j_{x_p}{x_p}_i+\mu^j_{x_q}{x_q}_i]\right.  \left. -\sqrt{\mu^j_{x_p}\mu^j_{x_q}}[\overline{\mu}^j_{x_p}{x_p}_i+\overline{\mu}^j_{x_q}{x_q}_i]\right) 
\label{eq:reg_rmb_update}
\end{aligned}
\end{equation}
where $\overline{\mu} = 1-\mu$, the $1/2$ is absorbed in the regularization trade-off parameter $\alpha$, and $x^{(k)}$ and $h^{(k)}$ are obtained with $k$-step Gibbs sampling.

The DR does not require any constraints on the visible units, and it can be used in combination with real-valued Gaussian-Bernoulli RBMs as well. 

\subsection{Variational Autoencoders (VAEs) Using Side Information}
Variational Autoencoders (VAEs) \cite{kingma2013auto,doersch2016tutorial} represent data generation process by an unobserved continuous latent random variable $H$ which is decoded with neural networks. The parameters of VAEs are trained jointly by backpropagation algorithm via optimizing the following cost function:

\begin{equation}
\begin{aligned}
\auxL(\theta,\phi;\D)= \sum_{\x\in\D}\E_{Q_\x}[\log P(\x|H;\theta)] - \sum_{\x\in\D}\Diverg{KL}{Q_\x^\phi}{P(H)}.
\label{eq:VAE_cost}
\end{aligned}
\end{equation}

The first term is the expected reconstruction error, while the second term, the KL-term,  pushes the posterior $Q$ toward the prior $P(H)$. We note that $Q$ tends to ``be covered'' by $P$ in this case. 





The DR can be introduced over the reconstruction level and we can have several options. For instance, DR could be defined via cross-entropy as follows:
\begin{equation}
D_{CE}(\x_p,\x_q)=\E_{Q_{\x_q}}[\log P(\x_p|H;\theta)],
\label{eq:VAE_RE_DR}
\end{equation}
However, when one does not require a probabilistic interpretation over the last layer, simply an Euclidean distance could be used as:
$D_2(h_{\x_p},h_{\x_q})=\|h_{\x_p}-h_{\x_q}\|_2^2,$
where $h_{\x_a}$ denotes the reconstruction of data $\x_a$ obtained with VAE and 2 in $D_2$ indicates its relation to the squared $L_2$-norm.


\section{Diversifying Regularization Based on Side Information for Discriminative Deep Models} 

In the case of DNNs, let $\h_{\x_p}^l$ denote the features obtained at hidden layer $l$  by forward propagation from data $\x_p$. Because $\h_{\x_p}^l$ does not have any probabilistic interpretation, we can define DR for two different types of data $\x_p$ and $\x_q$ by any distance function. Here, we define it as $D_2(\h_{\x_p}^l,\h_{\x_q}^l)= \|\h_{\x_p}^l-\h_{\x_q}^l\|_2^2$. Therefore, the learning objective including $D_2$ can be formulated as follows:
\begin{equation}
\begin{aligned}
J(\theta;\D) = \sum_{\x\in\ D} \ell(G(\x;\theta),y) + \alpha \Omega(\theta)  -\sum_{(\x_p,\x_q)\in \N}\sum_{l=1}^{L-1} \alpha_l D_2(\h_{\x_p}^l,\h_{\x_q}^l),
\label{eq:discriminative_cost}
\end{aligned}
\end{equation}
where $\ell$ is a loss function such as cross entropy, $G(\x;\theta)$ denotes the output prediction by DNN,  $\Omega$ is a parameter regularization, and $\alpha$s are trade-off values. The derivatives of this learning objective can be calculated analytically; however, it can also be done automatically via, for instance, Theano's grad function.

One can apply different trade-off coefficients $\alpha_l$ at different layers. Perhaps, at lower level feature learning, it is not expected to obtain different representations for different types of data; however, at higher levels, closer to the prediction level, stronger regularization $\alpha_l$ would be desired. 

\section{Experimental Results and Discussion}
The aim of our experiments is to demonstrate that training algorithms achieve better generalization performance with using DR than without using it under exactly the same parameter settings. We did not use any sophisticated methods such as adaptive learning rate, momentum methods, second order optimization methods, data augmentation, image preprocessing methods, etc, which would make it more difficult to reveal the contribution of DR to the final result. We note that,  therefore, our classification results do not surpass the best ones in the literature.  Here, we present three types of experiments, one for RBMs, one for VAEs, and the other for discriminative models. All the samples for the negative pairs were constructed from all the data having different class labels in the current mini batch.

\paragraph{On the Generative Learning of Deep Belief Networks}

First, we built a deep belief network (DBNs) \cite{larochelle2009exploring} on the Cifar-10 dataset \cite{krizhevsky2009learning}. The dataset contains tiny $(32\times 32)$ color pictures from 10 classes, 50,000 for training and 10,000 for testing. We constructed a DBN of  11 layers, each having 500, 300, 200, 150, 100, 80, 60, 50, 30, 20 hidden and 10 output units, respectively. The weights in every layer were pre-trained by an RBM using learning rate  0.01, one-step CD $(k=1)$. The batch-size was set to 10. For DR, we used the formula Eq. \ref{eq:RBMDR}, and all data in the mini-batch were used to construct the side information $(N)$, which resulted in approximately 35-45 pairs in each batch. The $\alpha$ trade-off parameter of DR was set to 50. For the fine-tuning backpropagation, we used a stochastic gradient descent without any parameter regularization (such as $L_2$), with learning rate 0.01 and batch size 10. The fine-tuning phases did not employ DR; therefore, we can see the effect of DR on the training of RBMs. All other parameters were left default. Before  training of both models (with and without DR), all weight parameters were initialized with exactly the same values. The DBN and RBM codes were downloaded from deeplearning.net.

\begin{figure}[tbp]
	\centering
	\small
	\includegraphics[width=13cm,height=6cm]{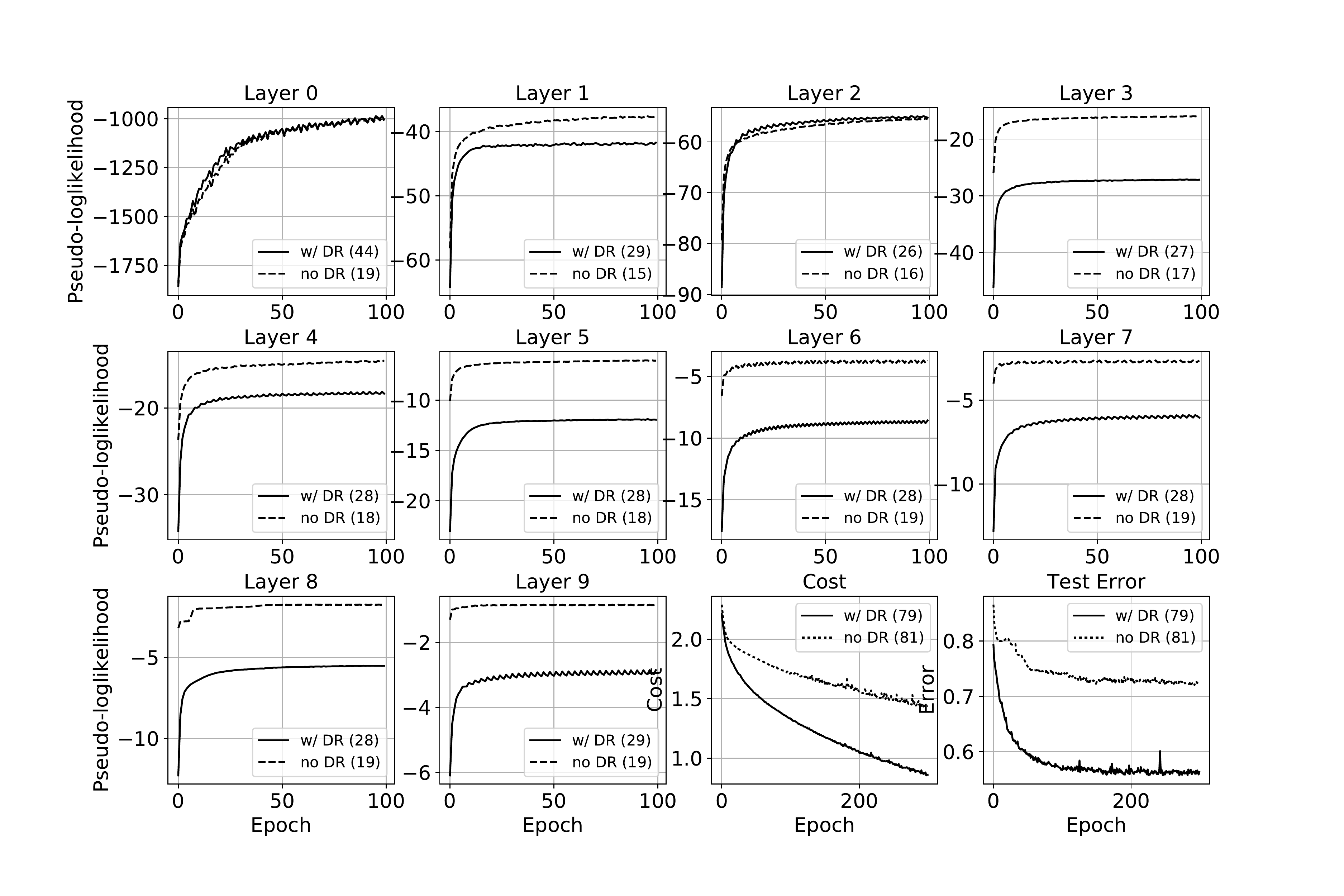} 
	\vspace{-0.5cm}	
	\caption{ Learning curves during training of RBM and DBN. The numbers in the parenthesis in the legends indicate the run time in minutes. Cost is defined as cross entropy.}
	\label{fig:Learning_curves_of_DBN}
	\vspace{-0.5cm}	
\end{figure}

The results are shown in Figure \ref{fig:Learning_curves_of_DBN}. The first 10 plots (first five rows) show the pseudo-log likelihoods obtained during training RBM with DR (solid) and without DR (dashed) at every layer. The last two plots in the last row show the cost and the test error during the fine-tuning, when it was applied after regularized (solid) and unregularized (dashed) pre-training. These plots show that the fine-tuning achieved much faster convergence and much better generalization performance when the weights were pre-trained using DR. We explain this as being due to the fact that DR includes class information about the data, which might not maximize the pure log likelihood but favors solutions where different types of data have more different abstract representations as well. Run times are indicated in the figure legend. Training using DR required more time because side information was generated on the CPU for each mini-batch inline. This possibly could have been accelerated by generating it for each batch before the training procedure, in advance.


\paragraph{On the Generative Learning of VAE}
Next, we examined the impact of DR, defined as in Eq. \ref{eq:VAE_RE_DR}, on training of VAE. We employed the classic MNIST dataset \cite{lecun1998gradient}, which contains $(28\times 28)$ gray-scaled pictures of handwritten digits: 50,000 of them for training and 10,000 for testing. The VAE consisted one hidden layer with 600 hidden units on the encoding and decoding layer, respectively, while the latent space was 2 dimensional. The training was carried out using the reparameterization trick \cite{kingma2013auto}. The prior and the posterior were defined as $P(H)=N(0,I)$ and $Q_\x^\phi=N(z|\mu(x), \sigma^2(x)I)$, respectively, where $\mu=\mu(x)$ and $\sigma=\sigma^2(x)$ are the outputs of the encoding neural network trained on data. Training run 100 epochs with learning rate 0.05 and batch-size 20.

\begin{figure}[!t]
	\centering
	\footnotesize
	\setlength{\tabcolsep}{0pt}	
	\begin{tabular}{cc}
		A)\raisebox{-0.9\height}{\includegraphics[height=3.5cm]{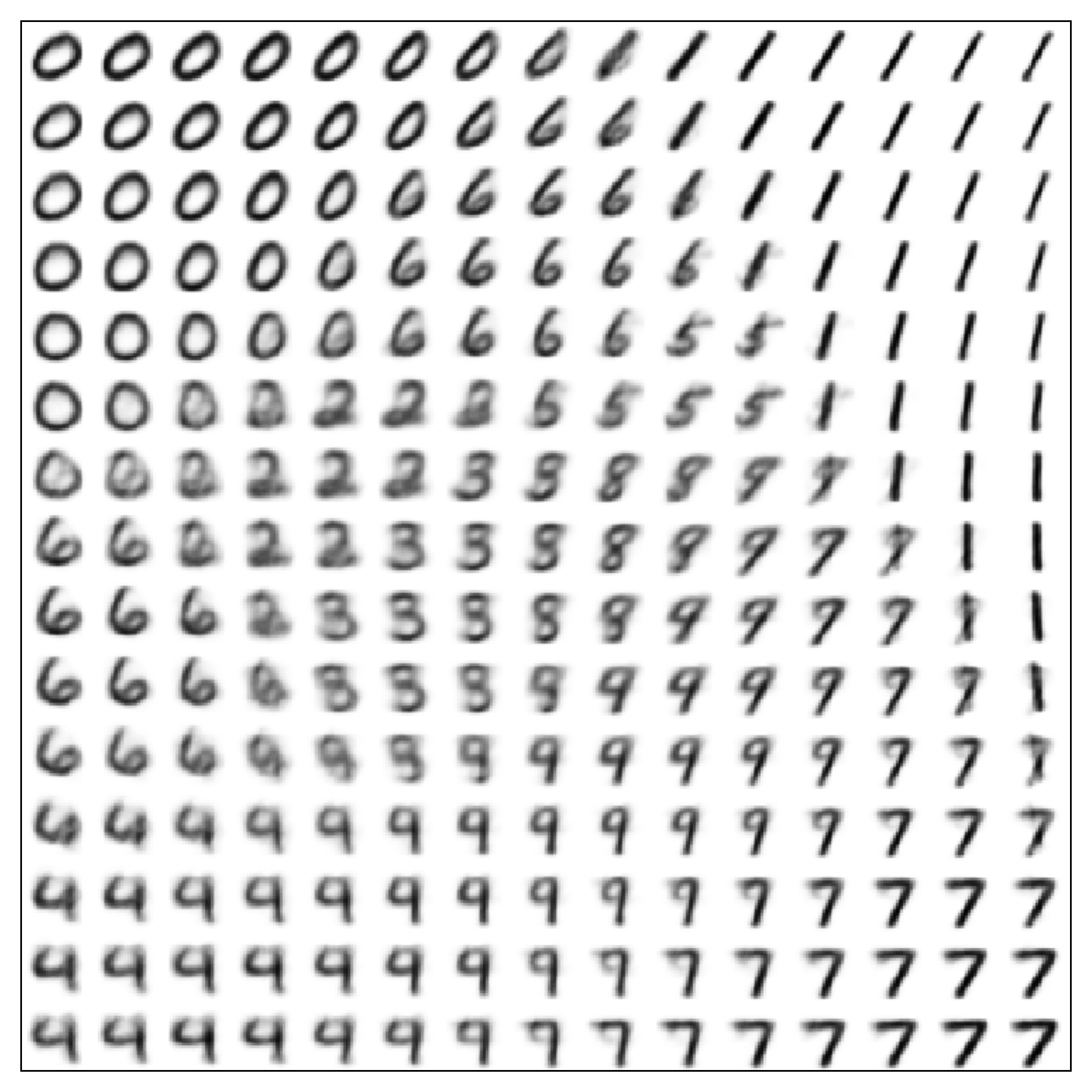}} & B)\raisebox{-0.9\height}{\includegraphics[height=3.5cm]{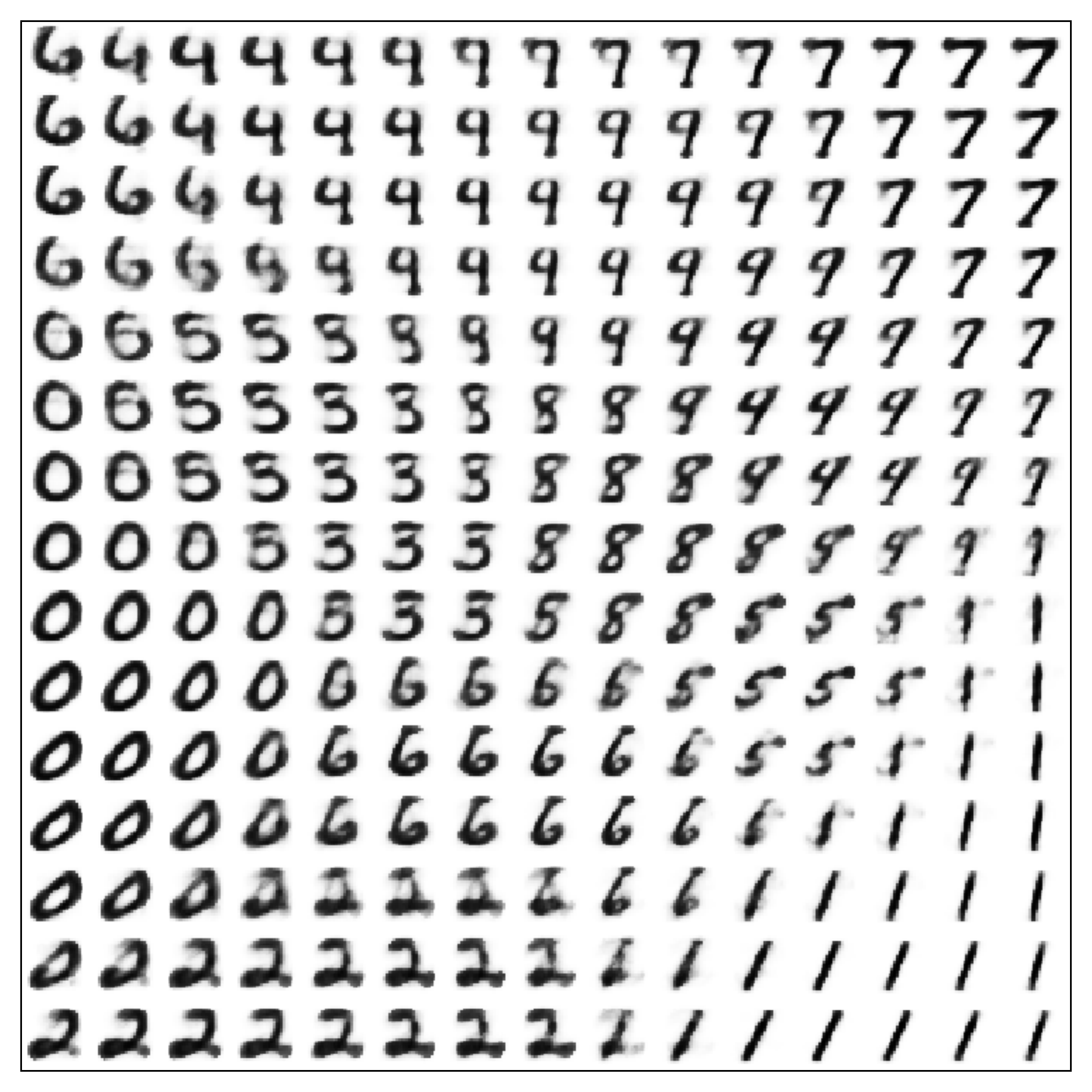}} \\
	\end{tabular}
	\caption{ Learned MNIST manifold (A) without regularization and (B) using DR. DR helps generate sharper images.}
	\label{fig:Learned_manifolds_of_VAE}
	\vspace{-0.5cm}
\end{figure}
For the visualizations of the learned manifolds, the linearly spaced coordinates of the $[-6,-6]\times[6,6]$ square were used as the latent variables $H$. For each of these values we plotted the corresponding generative $P(x|H)$ with the learned parameters. The sampled images are shown in Figure \ref{fig:Learned_manifolds_of_VAE}. The panel (A) shows the sampled images from the VAE trained without DR, while panel (B) shows sampled images from the VAE trained using DR. In our opinion, the decoder layer of the VAE trained using DR can produce sharper images.

\paragraph{On the Discriminative Learning of DNN}
Finally, we examined the impact of DR, defined in Eq. \ref{eq:discriminative_cost}, on training deep, fully connected sigmoid networks. We employed the MNIST dataset. We constructed a six-layer DNN, with hidden layers containing 30, 30, 30, 20, 20 units and the output layer containing 10 units. Training was carried out via a gradient descent algorithm without any parameter regularization (such as $L_2$), with learning rate 1.0. The $\alpha$ trade-off parameter of DR was set to 50.0, and it was reduced in each epoch by 10\%; therefore, its effect was strong at the beginning of the training but over the time it gradually vanished. For side information, 202,770 data pairs were randomly chosen at the beginning of the training.  Before training of both models (with and without DR), all weight parameters were initialized with exactly the same values. 

\begin{figure}[!htbp]
	\vspace{-0.5cm}
	\centering
	\small
	\includegraphics[width=9cm]{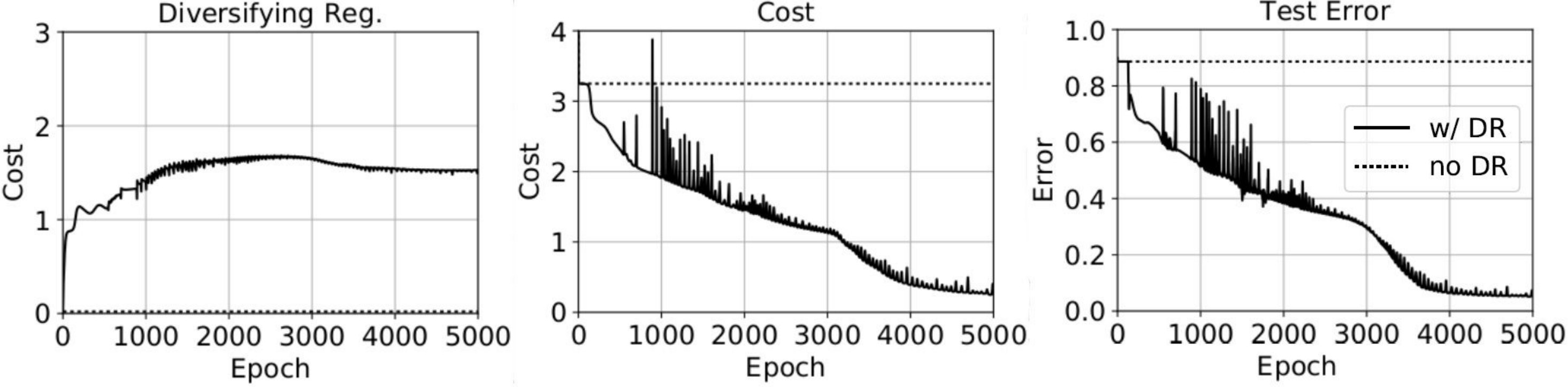} \\
	\vspace{-0.5cm}
	
	\caption{ Curves during training fully connected sigmoid DNN. Cost is defined as cross entropy. Middle plot does not include DR.}
	\label{fig:Learning_curves_of_DNN}
	\vspace{-0.5cm}
\end{figure}
The results and the learning curves are shown in Figure \ref{fig:Learning_curves_of_DNN}. The plots show the value of the regularization, the value of the cost function (without regularization) during training, and the error obtained on the test data, respectively. These plots show that the standard backpropagation algorithm (dashed line) was not able to train the model on this structure, and it gave an 88.65\% classification error.  We note that the tricky part here is the first hidden layer because the information compression is very high (784 $\rightarrow$ 30), so the backpropagation must train the corresponding weights very well. However, in practice the backpropagation failed to propagate the error from the output layer to the first layer.However, DR  (solid line) is applied directly on every layer providing useful class-label-related information locally. The backpropagation method using DR (solid line) during the first 127 epochs did not change the cost and the test error but it rapidly increased the diversification until it reached a critical point from where the gradients of the cost function led to a good (local) minimum. This resulted in a 5.44\% classification error.

\section{Conclusions}
In this article we have introduced a novel regularization method, termed diversifying regularization (DR), to help train deep generative and discriminative models for classification tasks. The main idea behind DR is that different types of data should have different abstract representation in the model's hidden layers. Therefore, DR applies a penalty on hidden units if they obtain similar features on data belonging to different classes. In the experimental results on deep belief networks, we have shown that DR is capable of including essential information about class labels implicitly and propagating them through layers in the greedy layer-wise pretraining. Therefore, the regularized weight initialization already includes some information about the class labels, and the subsequent fine-tuning phase can obtain smaller generalization errors and faster convergence. On the other hand, discriminative training of deep models can also benefit from DR, because it can provide good gradients at low layers, directly helping coping with vanishing gradient problem. 

Our method, DR, is not limited to DNNs, VAEs, RBMs, and DBNs. It can be included in many generative models and methods, such as Gaussian mixture models and wake-sleep algorithms as well.


\paragraph{Acknowledgments}
We gratefully acknowledge the support of NVIDIA Corporation with the donation of the GTX Titan X GPU used for this research.


\bibliographystyle{spmpsci}
\bibliography{bibliography}

\end{document}